\documentclass[runningheads]{llncs}
\usepackage[T1]{fontenc}

\usepackage{epsfig} %
\usepackage{mathptmx} %
\usepackage{times} %
\usepackage{xspace}
\usepackage{amsmath} %
\usepackage{amssymb}  %
\usepackage{multirow}
\usepackage{subcaption}
\usepackage{booktabs}
\usepackage{lipsum}
\usepackage[export]{adjustbox}
\usepackage{wrapfig}
\usepackage{tikz}

\usepackage{graphicx}
\usepackage{hyperref}
\usepackage{xcolor}

\DeclareMathOperator{\similarity}{sim}
\newcommand{\eg}{\textit{e.g.}\xspace}
\newcommand{\ie}{\textit{i.e.}\xspace}
\newcommand{\z}{\mathbf{z}\xspace}
\newcommand{\id}{\textit{id}\xspace}
\newcommand{\ood}{\textit{ood}\xspace}

\newcommand{\mnd}{$\mathit{M}_{ND}$\xspace}

\newcommand{\modd}{$\mathit{M}_{OD}$\xspace}
\newcommand{\softm}{\textit{softmax}\xspace}
\newcommand{\x}{\mathbf{x}}
\newcommand{\ours}{\textsc{iConP}\xspace}
\newcommand{\bce}{\textsc{iBCE}\xspace}

\newcommand{\ond}{\textit{OND}\xspace}

\newcommand{\myparagraph}[1]{\textbf{#1}}

\newcommand*\circled[1]{\tikz[baseline=(char.base)]{
            \node[shape=circle,draw,inner sep=1pt] (char) {#1};}}

\title{Incremental Object-Based Novelty Detection with Feedback Loop}

\author{Simone Caldarella\inst{1}\thanks{Work mostly done during an internship at Toyota Motor Europe } \and 
Elisa Ricci\inst{1, 2} \and
Rahaf Aljundi\inst{3} 
}

\institute{University of Trento \and Fondazione Bruno Kessler  \and Toyota Motor Europe}

\begin{document}

\maketitle

\begin{abstract}

Novelty Detection tackles the problem of identifying samples that do not belong to classes (\id) observed by a given model during its training phase. 
In particular, \emph{Object-based Novelty Detection} (\ond) introduces an additional granularity compared to image-level novelty detection, allowing to process multiple objects at the same time.
The majority of novelty detection approaches either assume full control over the initial (pre)training phase, or solely focus on post-processing of the pretrained model outputs at inference time.
As a consequence, these solutions inevitably discard the precious knowledge within out-of-distribution (\ood) data encountered after deployment. 
In this work, we assume that feedback in form of cheap annotations, confirming or rejecting the model's novelty decisions on detected objects, can be incrementally obtained.  We then investigate how incorporating this feedback can strengthen models robustness. 
We propose both \circled{1} a  novel setting and \circled{2} benchmark for \ond with incremental feedback loop, focusing on the continuous incorporation of \ood information over the model's life-cycle. 
Furthermore, we propose \circled{3} a new method to effectively incorporate such feedback.
Our approach, named \ours, consists of a lightweight novelty detection module, optimized continuously with the received feedback.
Our method reduces the False Positive Rate (FPR) of the object detection model by $\sim$64\%, when leveraging the continuously received feedback, while maintaining the object detection performance on known classes unaltered.

\end{abstract}

\section{Introduction}
\label{introduction}

Novelty Detection, also referred to as Out-of-Distribution detection, falls under the extensive research domain of anomaly detection and focuses on the challenge of distinguishing instances of \emph{novel} classes, \ie, \ood samples, from instances of classes provided to a neural network during its training phase, \ie, in-distribution (\id) samples. 
This task is vital for any model deployed in a real environment as it would be impossible to define an a priori set of all possible objects in a dynamic continuously evolving world.
Differently from human reasoning, neural networks are--by design--not capable of distinguishing between \id and \ood instances at inference time and, even worse, they tend to make mistakes with high confidence, a problem recognized as \textit{overconfidence}~\cite{wei2022mitigating}. 
Many works have focused on normalizing the output probability scores or the logits before the \softm computation in order to mitigate overconfidence and improve novelty discrimination~\cite{hendrycks2017a,liu2020energy,hendrycks2019scaling}. 
While being computationally efficient, these solutions cannot solve the inherent overconfidence problem, leading to limited performance in real-life scenario. 
Recently, many novelty detection methods~\cite{hendrycks2019oe,liu2020energy,pei2022out,meinke2019towards,papadopoulos2021outlier,thulasidasan2021effective,ji2022proactive,azzalini2021minimally,chen2021atom,pmlr-v162-ming22a} have proposed to leverage auxiliary \ood data to improve the latent representations of \id and \ood distributions.
Particularly, a common winning trend was the limited exposure to a \textit{representative} set of randomly collected \ood samples, which resulted in a steady performance improvement, compared to \id-only solutions. For instance, Outlier Exposure \cite{hendrycks2019oe} integrates the \id vs \ood detection task, with exposed \ood data, as a regularizer  of the main model.
Regardless the application domain, all existing works share a crucial weakness: they all assume a one step training of the novelty detection method disregarding the abundant \ood data observed during deployment time.
We argue that these assumptions are highly limiting in many real-world applications, calling for novel solutions.

\begin{figure}[t]
    \centering
    \includegraphics[width=0.9\textwidth]{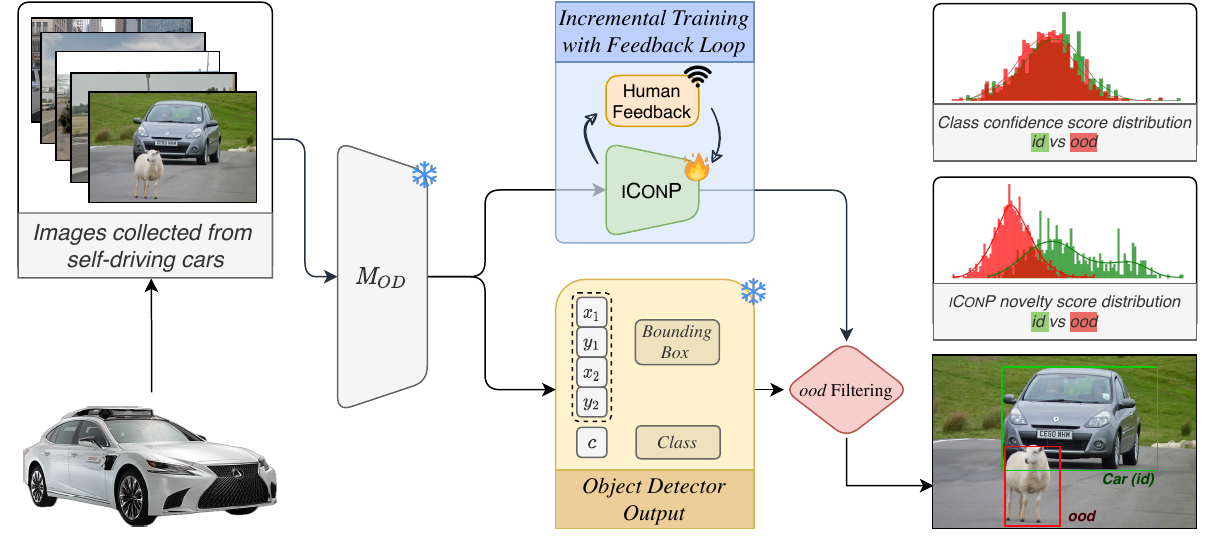}
    \caption{\footnotesize Our proposed 
    setting for \textit{incremental} object-based novelty detection \textit{with feedback loop}. During deployment, a novelty detection function (ours is coined \ours) is responsible for detecting whether a feature of an instance, extracted by the object detection model represents  an \ood or an \id object. These predictions, along with original pictures, are shared on the cloud with human annotators that provide weak labels in form of rejecting or accepting the model {\id}/{\ood} decisions. This feedback is then used by the model to increase its \ood robustness. Comparison of predicted scores distribution with Max Logit (baseline - no feedback loop) and \ours - ours after few sessions of feedback are shown on the right.}
    \label{fig:teaser}%
\end{figure}

To delve deeper into the problem, imagine a fleet of autonomous cars equipped with object detection models capable of recognizing specific classes like various types of vehicles, pedestrians, and traffic signs. Given the unlimited set of objects on the road, it is crucial to identify whether an object belongs to the classes of interest to avoid misleading predictions and unsafe actions. Due to the importance of distinguishing between in-distribution and out-of-distribution samples, predictions can be sent to the cloud for verification or correction by human annotators. This feedback, based on weak annotations (no class-level labels or bounding boxes needed), can improve the robustness to unseen out-of-distribution data and the reliability of the novelty detection method. \textbf{ How to best leverage these new annotations is the primary research question we address in this paper.}
Motivated by the scenario described, we focus on the underexplored problem of object-level novelty detection, which is crucial for real-life applications. Our approach is applicable to \textbf{any pretrained model} without prior knowledge.
We assume a pretrained object detection model that discriminates well between predefined objects and aim to enhance its robustness to out-of-distribution data using test-time feedback.  %
Specifically, to mimic a realistic setting, we assume that feedback is received at disjoint sequential sessions upon availability, and the object-based novelty detection has to be improved after each feedback session, creating a loop of deployment and training on recieved feedback (see Figure~\ref{fig:teaser}). %

For the \id vs \ood detection task, we  propose to independently optimize a lightweight novelty detection module on top of the (pre)trained object detection model.
Our specialized module uses the latent representations of detected objects and inspect whether those samples belong to the set of known classes or not. 
This design ensures efficient, low-cost updates to the module while maintaining the initial performance of the pretrained object detection model, improving system's robustness. 
As object detection models grow in scale and size, improving test-time robustness without costly retraining becomes crucial.

Recently, \cite{pmlr-v162-bitterwolf22a} showed that when representative \ood data can be leveraged, a binary classifier trained on the binary \id vs \ood problem is the optimal Bayesian function. %
Based on this finding, we treat the \ood detection as a binary discrimination problem between \id and \ood data. 
However, we observe that the naive solution of training the specialized novelty detection module with a Binary Cross Entropy (BCE) loss exhibits poor generalization capabilities, especially in an incremental scenario, and shows small unsteady improvements from the received feedback. %
To overcome this issue, we propose to leverage more powerful representation learning methods~\cite{balestriero2023cookbook} that operate on the sample pairwise similarities. %
Specifically, we learn the hidden representations of the novelty detection module using a Supervised Contrastive Loss~\cite{khosla2020supervised}, to group \id instances further from their \ood counterparts. %
To obtain a powerful novelty detector, we simultaneously optimize a 1D projection of the trained representation using BCE.
Our experimental evaluation shows that \ours is able to consistently improve generalization to unseen \ood classes in the incremental  setting, largely benefiting from the provided human feedback.
In order to evaluate our method in both the \textit{incremental} and the classical-\textit{static} setting, we introduce a novel benchmark, which comprises both evaluation protocols. Moreover, during our preliminary experiments,  we noticed that existing benchmarks consider \id and \ood instances of different datasets, leading to fake robust novelty detection performance where mostly domain shift is leveraged. More importantly, the dataset from which the \id samples  are drawn during evaluation, is commonly the same used for training the backbone model, i.e., the object detector in our setting.
We believe that this violates the test condition where \id and \ood are both sampled from the very same underlying distribution, different from the one seen during training. Our benchmark is built by sampling both \id and \ood data at test-time from the same test domain, which is different from the one used in the pretraining stage.

\textbf{Contributions.} Our contributions are as follows: 1) we formulate the novel problem of incremental object-based novelty detection (\ond) with feedback loop and 2) we introduce a novel approach based on a lightweight novelty detection module which shows robust performance and steady improvements from the received human feedback. To rigorously evaluate our approach we propose a new benchmark which overcomes the limitations of previous works~\cite{du2022vos,hendrycks2017a,liang2017enhancing}. 3) We show that existing novelty detection approaches suffer from severe performance drop when \id and \ood data come from the same testing distribution and from a different distribution than the one used for training the backbone model, and that our approach can effectively increase the robustness of the deployed object detection model  as feedback is received.

\section{Related Works}
\label{relatedworks}
\textbf{Novelty Detection.} Existing approaches for novelty detection can be roughly grouped in two categories: post processing and regularization. 
Earlier works mostly belong to the first category and 
they operate on the model output logits or intermediate features. For instance, \cite{hendrycks2017a} proposed MSP, a simple yet intuitive baseline to perform novelty detection using the predicted class probability (\ie{}, maximum softmax probability) as an \id{} score. 
Differently, energy-based OOD detection~\cite{liu2020energy} deploys energy scores which are theoretically aligned with the probability density of the inputs and less susceptible to the overconfidence predictions  of MSP.  
 Later, Max Logit~\cite{hendrycks2019scaling}, the maximum logit value before the \softm{} function, was proposed as a robust detection function in large scale and realistic outlier detection problems. 
Differently, regularization-based methods modify the model enforcing more diverse predictions for \id{} vs \ood{} samples and typically involve (re-)training or fine-tuning of the main model. \cite{inproceedings} proposed an unsupervised approach considering the reconstruction error from a learned autoencoder as novelty detection score. 
\cite{Dwibedi_2021_ICCV} introduced a framework based on contrastive learning and proposed to improve the quality of the learned latent representations using nearest-neighbours of samples as positive pairs. \cite{tack2020csi} use strongly shifted instances as a proxy for \ood{} data and adopted contrastive learning where, during training, each sample is contrasted with other instances and with its distributionally-shifted augmentations. 
Similar to these latter approaches our method is also based on contrastive learning. Nonetheless our target here is different from previous methods, as we aim to learn a projection in order for \id{} samples to lie close together in the embedding space and far apart from \ood{} samples.
Majority of image-based Out-of-Distribution detection approaches, classify the full image as \id{} or \ood{}, recently VOS~\cite{du2022vos} has emphasised  the importance of outlier detection for object detection models. VOS proposed to synthesize outliers in the representation space and train a multi-layer perceptron (MLP) to maximize the energy of \id{} data while minimizing it for \ood{} data. Our solution,  
differently, involves training an MLP to directly separate \id{} from \ood{} representations. 

The above works assume that access is available only to \id{} data. Nevertheless, given that \ood{} examples can be gathered, new methods have been proposed to train directly on \id{} and \ood{} data for better detections.
Outlier Exposure~\cite{hendrycks2019oe} proposes to combine Cross Entropy (CE) loss minimization on \id{} data with another CE loss on available \ood{} data, here with uniform distribution as a target instead of the one-hot class labels.
\cite{liu2020energy} suggested alternatively to maximize the energy on \ood{} data while minimizing it for \id{} data. These approaches use an auxiliary \ood{} dataset of million images. 
More recently ~\cite{seifi2024ood} propose to extend supervised contrastive learning~\cite{khosla2020supervised} with \ood related loss terms to encourage \ood{} samples to be projected further from the clusters of \id samples. The method, however, tunes or pretrains the whole classification model. Differently, Atom~\cite{chen2021atom}  proposes an approach which mines important auxiliary \ood{} data, however in one offline step.
Given the increased amount of works leveraging \ood{} data, \cite{pmlr-v162-bitterwolf22a} analyse common ND methods, and theoretically demonstrate that novelty detection can be regarded as a binary discrimination problem between \id{} and \ood{} data. Inspired by \cite{pmlr-v162-bitterwolf22a} we approach outlier detection as a binary classification problem. However, we show that simply training a binary classifier using binary cross entropy is not an optimal solution. 

\textbf{Continual and Active Learning.} Our work shares various similarities with existing literature on continual and active learning. Continual (incremental) learning methods~\cite{de2021continual} attempt to sequentially learn a model for a task without forgetting knowledge obtained from the previous tasks. Common lines of work can be loosely divided into regularization approaches~\cite{mas,ewc,lwf} and rehearsal based approaches \cite{ER,DER}.
Similarly, in this paper we also learn \emph{continually} to distinguish between \id{} and \ood{}, assuming \emph{continuous} observations and updates of the \ood{} detection model to further improve its robustness to novel unseen \ood{} classes. Inspired by  GDumb~\cite{prabhu2020gdumb}, we do not employ a regularization technique to avoid catastrophic forgetting but simply rehearse all the stored features. As mentioned in~\cite{prabhu2020gdumb}, storing all the required features (in our case obtained from the observed samples through the object detector) has a negligible impact on the storage cost  of the entire pipeline (e.g., $\sim$4kb each feature vector) which can be lighter than a parameter regularizer with a storage and memory cost equal to a model size of parameters . %
This allow us to seamlessly update the model with a minimal expense (see \ref{compover}).

On the combination of continual learning and novelty detection, one closely related work is Continual Novelty Detection \cite{pmlr-v199-aljundi22a}, that investigates the performance of ND methods when coupled with continually learned models where the set of \id{} classes changes over time. Differently, in  our setting we assume a fixed \id{} set whilst obtaining incrementally feedback on new \ood{} classes.
Closely,~\cite{rios2022incdfm} move a step forward and study the setting where an \ood{} detector is used to detect \ood{} samples which are then modeled as new ID classes and learned continually. Our approach assumes that the set of concerns \id{} classes are fixed and aim at strengthening the robustness of an \ood{} detector as new \ood{}/\id{} samples are encountered and feedback is received.
Our proposed setting share some similarities with active learning~\cite{ren2021survey}, as we also invoke the supervision of an oracle from the cloud. However, the emphasis in active learning methods is on estimating confidence for choosing the appropriate samples and quickly improve the predictor, while \ours{} focus on improving robustness to novel and unseen sample classes.

\section{Proposed Setting}
\label{setting}
In the proposed incremental object-based novelty detection (\ond) with feedback loop we consider a pretrained object detection model \modd{}. Several existing object detectors are currently available, such as Faster-RCNN~\cite{frcnn}, YOLO~\cite{yolo}, DETR~\cite{detr} or even more recent open world object detection models such as Grounding DINO~\cite{liu2023grounding} and Yolo-World~\cite{cheng2024yolow}.

The object detection model is pretrained on large scale data and is deployed to detect instances of a predefined set of objects. This set is referred to as In-Distribution: $\id=\{(img,~b,~y):~y~\in~\{c_1,~..,~c_K\}\}$ where $img$ is a given image, $b$ is a bounding box, $y$ is the corresponding class label and $K$ is the number of the classes the model has to detect. 
The object detection pipeline is equipped with a novelty detection model \mnd{} used to reject detections of object instances that belong to  classes not contained in the \id set: $\ood=\{(img,~b,~y):~y~\notin~\{c_1,~..,~c_K\}\}$. Typically, \mnd{} predicts an \id{} score that is high for \id{} samples and low for novel unknown objects (\ood{} samples).

Our setting is illustrated in Fig~\ref{fig:teaser}. The target is to improve \mnd{} ability of identifying \ood{} instances, \eg, unknown type of vehicle on the road, an unknown object in a hallway, etc. First, we start with an initial set containing both \id{} and \ood{} samples gathered from the same deployment environment as where the model is operating. This set is used to refine the novelty detection to the characteristics of the deployment environment, \eg, a given city or a given factory. We refer to this training session as $S_0$. Afterwards, the model is deployed and it has to recognize encountered objects that belong to the \id classes, whilst filtering objects of unknown classes. These decisions are then sent to the cloud where they are inspected by a human annotator and labelled according to their membership either to known or unknown set of classes. The collected annotations are then used to retrain the specialized novelty detection module \mnd{} during the next session $S_1$ in order to refine its detection capabilities. This procedure can continue as long as mistakes are made and annotation budget exists. It is worth noting that the newly annotated \ood{} data can belong to other unknown classes than what is encountered in previous training sessions. The target is to improve the \ood{} detection of any possible unknown object especially those that have never been encountered before, \ie, \textit{unseen}-\ood{} classes. Importantly, the main object detection performance by \modd{} on \id{} instances   should not be negatively affected.%

\section{Method}
\label{method}
\begin{figure}[t]
    \centering
    \includegraphics[width=1\textwidth]{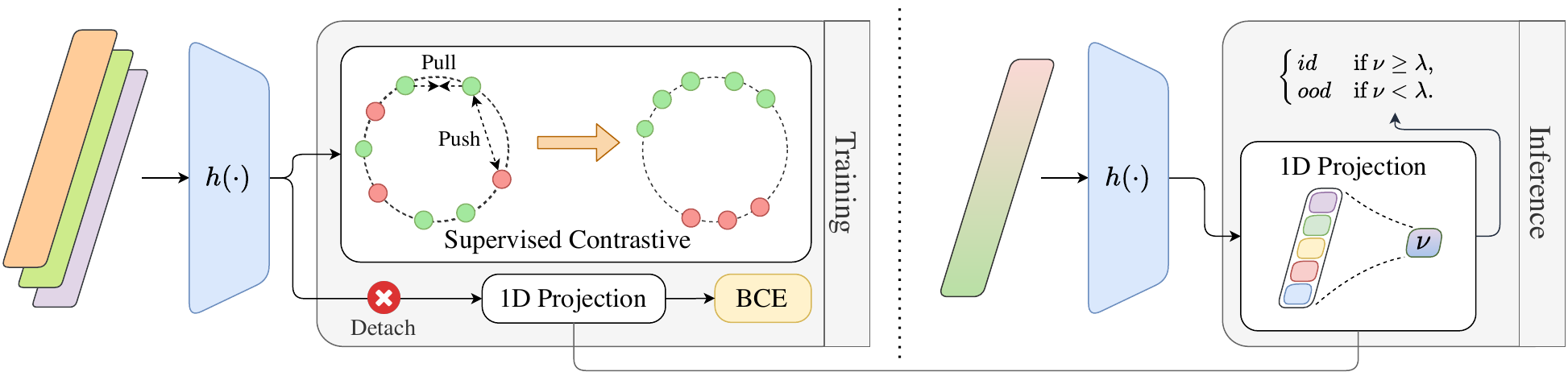}
    \caption{\footnotesize Our proposed  approach. During training (left) we optimize a light-weight ND module to project features from a pretrained object detection model into a hyper sphere, where \id{} instances are grouped together and far from \ood{} samples; jointly  a scoring function is  optimized as a 1D projection of learned latent features  with BCE. At test time (right) \id{} samples would be assigned high score indicating an instance of a known object as opposed to \ood{} samples.}
    \vspace{-0.6cm}
    \label{fig:method}
\end{figure}

\subsection{Overview}
Given an input image, the pretrained object detection network \modd{} extracts representations of object proposals $\{\x_i\}_{i=1}^{P}$ with their associated bounding boxes $b_i$. The representations $\{\x_i$\} are then fed to a lightweight network \mnd{}, parameterized with parameters $\phi$ and trained in order to predict the likelihood of a given proposal to be associated with \id{} classes rather than an unknown class. 
\mnd{} maps each proposal representation $\x_i$ to a scalar  representing the detection score $\nu_i=f_\phi(\x_i)$.
Training is performed on two sets, one containing proposals of \id{} objects and the other containing \ood{} objects proposals. %

\vspace{-3mm}
\subsection{The proposed approach}

Recently, \cite{pmlr-v162-bitterwolf22a} theoretically analysed different \ood{} training and scoring mechanisms when access to representative auxiliary data is available, and show that a Bayes optimal function of different novelty detection approaches is a binary discriminator of \id{} vs. \ood{} collected data. Based on this theoretical observation, in this work we treat the novelty detection task as a binary classification problem. One important distinction has to be made here. In the typical binary classification the goal is to generalize to unseen data of two specific classes, while in our novel setup, the \ood{} set representative of all the possible classes other than the \id{} ones and the goal is to improve the generalization not only to unseen samples but also to unseen samples of never encountered \ood{} classes.
Here, we propose a novelty detection module \mnd{} optimized to project \id{} data in a space linearly separable from all possible \ood{} data.

A straightforward solution to train \mnd{} is optimizing it directly with Binary Cross Entropy (BCE) loss. We hypothesize that such basic solution might provide a little flexibility in refining the decision boundaries after receiving a new feedback, as training with BCE would be driven by learning decision boundaries between two sets with no regard to the internal similarities among \id{} samples. 
On the other hand, a pairwise loss, such as Supervised Contrastive loss (SupCon)~\cite{khosla2020supervised}, would try to find a new representation where all \id{} samples lie close together and far from all other \ood{} data, and similarly for \ood samples. We suggest that finding such embedding rather than directly optimizing a discriminative decision boundary provides a basis for smoother \id{}-\ood{} distributions separation and more flexibility to later adjustments when new \ood{} classes are received for re-training. 
To this end, let $f$ represents the function approximated by \mnd{}, we express f as a composition of two functions $f(\x_i)=g(h(x_i))$ were $h$ maps $\x_i$ to a latent variable $\z_i$ and $g$ maps $\z_i$ to its corresponding novelty score $\nu_i$. We optimize the hidden representation $\z$ of \mnd{} %
using SupCon loss, as follows:
\begin{equation}
\ell_{SupCon}=\frac{1}{N}\sum_i^{N}\ell_{SupCon}(\z_i, P_i),
\end{equation}
~\vspace{-0.3cm}
where:
\begin{equation}\label{eq:supcon}
\ell_{SupCon}(\z_i, P_i)=
-\frac{1}{|P_i|}\sum_{\z_p\in P_i} \log\frac{\exp(\similarity(\z_i,\z_p)/\tau)}{\sum_{j\ne i}\exp(\similarity(\z_i,\z_j)/\tau)},
\end{equation}

and $\similarity(\z_i,\z_j) = \dfrac{\z_i^\top \z_j}{||\z_i||\cdot||\z_j|| }$, $P_i$ is the set of representations $\z_p$ forming positive pairs for the $i$-th sample (\ie, the representations of proposals of the same set of objects, \id{} or \ood{}), the index $j$ iterates over all samples, $N$ is the total number of \id{} and \ood{} samples, and $\tau$ is a temperature parameter.
We do not consider any data augmentation.

To estimate a novelty detection score, we propose to optimize a one dimensional projection $g$ that associates \id{} data with high scores as opposed to \ood{} data. We approach this by minimizing BCE loss on the hidden representation  ($\z_i$).
\begin{equation}
    \ell_{BCE}=-\frac{1}{N}\sum_i^{N}  \ell_{BCE}(\z_i, y_i),
\end{equation}
\begin{equation}\label{eq:bce-sub}
    \textit{where}\quad\quad\ell_{BCE}(\z_i, y_i)=- \left[ y_i \log(g(\z_i)) + (1-y_i) \log(1-g(\z_i))\right],
\end{equation}
\begin{equation}\label{eq:g}
\hspace{-2.7cm}
    \textit{and}\quad\quad\quad\quad\quad\quad\nu_i = g(\z_i)=\frac{1}{1+ e^{-(\mathbf{w}^\top st(\z_i))}},
\end{equation}
where $st$ denotes the stop gradient operation and $\mathbf{w}$ is the weight vector optimized for the 1D projection. Here, the binary cross entropy is only optimizing the final 1D projection to find the best separating direction, and it has no effect on the representations being trained. Our full training objective is: 
\begin{equation}\label{eq:nd-obj}
    \ell_{ND} = \ell_{SupCon} + \ell_{BCE}.
\end{equation}
\vspace{-1.1cm}

\subsection{Incremental Updates}
Our goal is to obtain a strong novelty detector that can be incrementally improved whenever feedback is received, without acting on the pretrained object detector. 
We attach our module \mnd~ to the pretrained object detector in order to receive the encoded object proposal features.
The first step of training \mnd{} is based on offline gathered \id{} and \ood{} data from the expected deployment distribution as we explained earlier. We optimize \mnd{} using \eqref{eq:nd-obj}. Then, when a new annotated \id{}-\ood{} set is received, we finetune \mnd{} initialized with previous session's parameters.

Directly finetuning the model on a new set of \ood{} classes will drift it from what it has learned before, causing forgetting of the previously captured decision boundaries~\cite{de2021continual}.  
Here to prevent forgetting, we store the representations $\{\x_i\}$  extracted by the OD model of the previously observed sets. We deploy Experience Replay~\cite{chaudhry2019continual} where representations of previous sets are replayed jointly with the new set samples. 
We do not use any explicit regularizer here to further control forgetting, as it was shown in~\cite{davari2022probing} that representation trained by Supervised Contrastive loss is much less prone to forgetting than alternatives based on cross entropy. We leave the investigation of how more effective continual learning methods can be combined with our framework to future work.

\section{Experiments}
\label{experiments}
We first explain our setting and benchmark construction. Then, we compare our proposed approach with different novelty detection methods.

\vspace{-2mm}
\subsection{Test-Domain Incremental Benchmark}
\label{subsec:tdib}
Novelty detection evaluation benchmarks are constructed such that \id classes are drawn from the same distribution as the training data for the pretrained model, while the \ood classes are drawn from a new distribution, \ie, a different dataset~\cite{du2022vos,hendrycks2017a,liang2017enhancing}. 
This setup is relatively easy as the novelty detection problem boils down to distinguishing samples from different datasets and that might result in a good but misleading performance. 
Some recent benchmarks for classification divide a given dataset into sets of known (\id) and novel (\ood) classes overcoming the dataset shift effect, however,  the test distribution of a given model is assumed to be the same as the training distribution.

To emulate a real-world scenario, we introduce a benchmark called the Test-Domain Incremental Benchmark. The process of creating this benchmark relies on the initial pretraining set solely as a reference to select \id classes and the key is to identify another dataset that contains large set of objects classes with \id classes as a subset. 
We refer to the pretraining (or tuning) dataset of the main object detection model as ${D}_{train}$, containing a set of \id classes ${C_{\id}} = \{c_1,~c_2, ~...,~c_K\}$. For evaluating the novelty detection performance, we consider  a test dataset ${D}_{test}$  that contains object classes ${C_{test}}=\{c_1, c_2, ...,~c_K,....,~c_{K+J}\}$ where the \id classes  ${C_{\id}}$  form a subset of  ${C_{test}}$.
\begin{table}[t]
\small
\centering
\caption{\footnotesize \ond performance on VOC-COCO Offline (Classical and Test Domain) benchmarks \textit{before the feedback loop}. Both \bce and \ours improve significantly over compared methods. VOS \cite{du2022vos} performs poorly on Test Domain benchmark.  }
\vspace{2mm}
\begin{tabular}{@{}lrrcrr@{}} \toprule
\multirow{2}{*}{\textbf{Method}} & \multicolumn{2}{c}{\textbf{VOC (ID) vs COCO (OOD)}} & \phantom{abc} & \multicolumn{2}{c}{\textbf{COCO (ID) vs COCO (OOD)}}\\
\cmidrule{2-3} \cmidrule{5-6}
& FPR@95$(\downarrow$) & AUROC ($\uparrow$) & & FPR@95 ($\downarrow$) & AUROC$(\uparrow$) \\
\midrule
MSP \cite{hendrycks2017a}             &         74.32 &          78.47 & &          81.62 &          74.12 \\
Energy OOD \cite{liu2020energy}       &         64.30 &          77.60 & &          70.14 &          75.97 \\
Max Logit \cite{hendrycks2019scaling} &         63.88 &          77.66 & &          69.62 &          75.96 \\
VOS \cite{du2022vos}                  &         49.11 &          87.54 & &          81.70 &          74.44 \\
\midrule
\bce                                  & \textbf{22.96} & \textbf{95.22} & & \textbf{44.78} &         88.96 \\
\ours                                 &         27.45  &         94.16  & &         45.37  & \textbf{89.90} \\
\bottomrule
\end{tabular}
\vspace{-3mm}
\label{tab:initial}
\end{table}

\vspace{-4mm}
\subsubsection{Incremental Steps}
\begin{wrapfigure}{r}{0.5\textwidth}
\vspace*{-16mm}
  \begin{center}
    \includegraphics[width=0.5\textwidth]{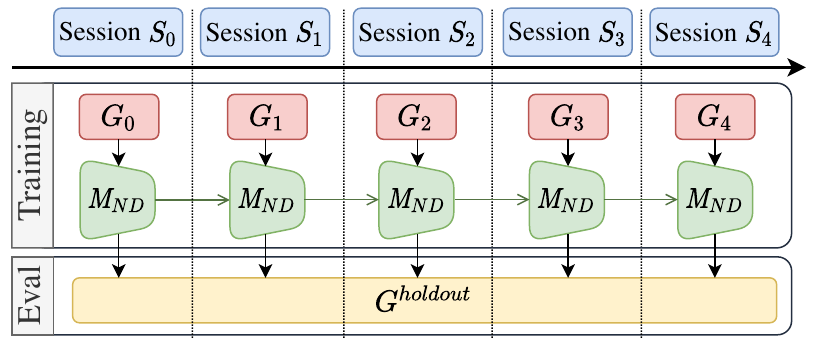}
  \end{center}
  \vspace{-4mm}
  \caption{\footnotesize Our incremental training and evaluation protocol. At each training session $S_i$ a new group of \id and \ood samples $G_i$ is received. Evaluation is performed on a distinct group of \id-\ood samples, namely $G^{holdout}$.}
  \label{fig:protocol}
  \vspace{-8mm}
\end{wrapfigure}

The target is to simulate the scenario where \mnd predictions are sent to the cloud and their \id-\ood annotations are received on which a new training session is conducted. We split the test dataset ${D}_{test}$ into disjoint groups of samples. Each group ${G_i}$ contains two subgroups, an \id subgroup $G^{\id}_i$ containing a unique set of images with instances of \id object classes ${C_{\id}} = \{c_1,~c_2,~...,~c_K\}$; and an \ood subgroup $G^{\ood}_i$ containing a unique set of images and objects instances of distinct \ood classes $C_{i}^{\ood} \in \{c_{K+1},~...,~c_{K+J}\} $. We leave one group $G^{\textit{holdout}}$ as a holdout set to evaluate the novelty detection performance on never seen set of images containing objects instances of \id classes and never seen \ood classes. Note that the constructed \ood subsets $\{G^{\ood}_{i}\}$ contain distinct set of classes.
The first group  ${G_0}$ is used in the first offline training session ($S_0$) of \mnd. At each later training session  ${S_i}$, \mnd is provided with  samples from ${G_i}$. Figure~\ref{fig:protocol} illustrates our training and evaluation protocol to strictly assessing the possible improvements in the generalization performance of \mnd  on a set of never seen \ood classes and never seen \id instances.

\subsection{Datasets}

We rely on three datasets. \textbf{VOC}, with over 10k images and 25k annotated objects across 20 classes \cite{pascal-voc-2012}; \textbf{COCO}, with 123k images and nearly 900k instances across 80 classes, including 20 VOC classes \cite{lin2014microsoft}; and \textbf{OpenImages}, the largest collection with 16M bounding boxes for 600 classes on 1.9M images, including all VOC and 60 COCO classes \cite{OpenImages}.

We construct two test benchmarks, one with ${D}_{train}$ as VOC and ${D}_{test}$ being COCO, where we train the object detection model on VOC. The second is with  ${D}_{train}$ as COCO and  ${D}_{test}$ being {OpenImages} imitating a large scale scenario. 
The ${D}_{test}$ in both cases is split into 6 independent groups $\{G_i\}$, corresponding to 5 training sessions $\{S_i\}$ plus the holdout group $G^{holdout}$ used for unseen \ood novelty detection evaluation. We refer to the Appendix A.1 for more details on the sets construction. %

\vspace{-3mm}
\subsection{Metrics}
As we train a separate novelty detection module and we don't alter in any way the original object detection model, but only exploit the extracted features, we focused on the metrics that are directly related to the task on novelty detection, ignoring those related to the detection task (mAP and IoU).\\
\textbf{FPR@95 ($\downarrow$)}, corresponds to the False Positive Rate when the True Positive Rate is set to 95\%. \\
\textbf{AUROC ($\uparrow$)}, the area under the Receiver Operating Characteristic (ROC) curve computed over all possible thresholds.%
\begin{figure}[t!]

    \centering
    \vspace{-13pt}
    \begin{subfigure}[t]{.49\textwidth}
    \centering
\includegraphics[width=1\textwidth]{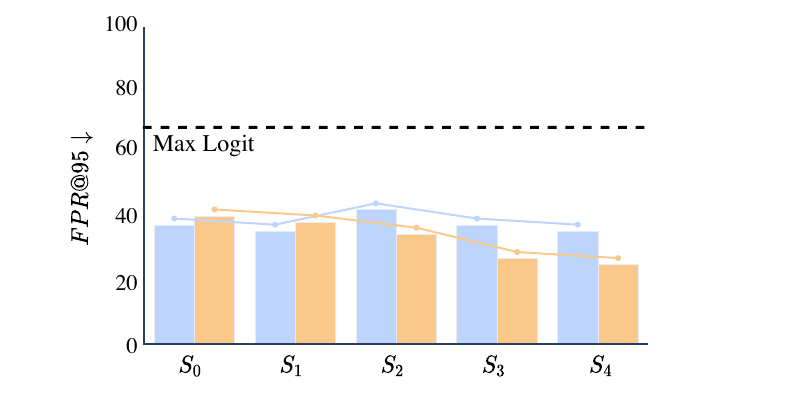}
        \vspace{-20pt}
        \caption{VOC-COCO Benchmark}
    \end{subfigure}
    \hfill
    \begin{subfigure}[t]{.49\textwidth}
    \centering
        \includegraphics[width=1\textwidth]{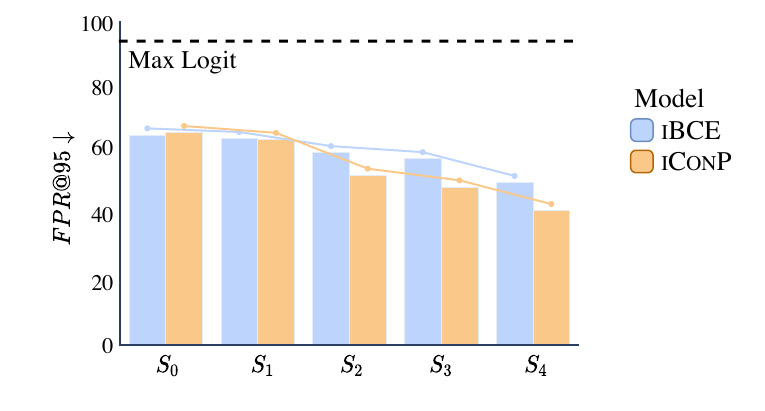}
        \vspace{-20pt}
        \caption{COCO-OpenImages Benchmark}
    \end{subfigure}
    \caption{Novelty detection performance (FPR@95, \textbf{lower is better}) of Max Logit (dashed line), \bce~and \ours~on the Test-Domain Incremental benchmark, evaluated after each session on $\mathit{G}^{holdout}$. \ours is capable of steady improvements as feedback is received contrary to \bce.}
    \label{fig:incremental_steps}
\end{figure}

\vspace{-3mm}
\subsection{Compared methods}

We consider the following novelty detection methods that can be directly applied to the object detection problem.\\ %
\noindent\textbf{MSP.}~\cite{hendrycks2017a} The maximum \softm probability predicted by the object detection model for a given object proposal is used as an in-distribution score. \\
\noindent\textbf{Energy OOD.}~\cite{liu2020energy} The energy of the logits distribution is used as an in-distribution score. \\
\noindent\textbf{Max Logit.}~\cite{hendrycks2019scaling} another variant to MSP, directly uses the maximum logit value, before \softm. \\
\noindent\textbf{VOS.}~\cite{du2022vos} represents the state-of-the-art for the \ond task. An MLP is trained on the energy values of \id proposals and synthetic \ood features, and used for detecting \ood instances.\\
For our framework of incremental \ond from feedback, apart from the best performing baseline in the classical setting, we consider  an additional baseline \textbf{\noindent \bce}: \mnd trained with Binary Cross Entropy (BCE) on the \id vs. \ood classification problem. 
Our  method is named \textbf{\ours}, where \mnd is optimized with the objective defined in~\eqref{eq:nd-obj}.

\subsection{Implementation Details}
For {\modd}, on VOC we employed a Faster R-CNN based on ResNet-50 trained on VOC 2007, and VOC 2012 and for COCO trained backbone we used Faster R-CNN based on ResNet-50 with a Feature Pyramid Network~\cite{fpn} trained on COCO 2017.
For {\mnd} we deploy an MLP of 4 hidden layers plus 1 layer for the final projection to one output logit, optimized with Adam~\cite{adam}, for \bce, and with SGD, for \ours.
Further details regarding the implementation are reported in the Appendix A.1.%

\subsection{Results}
\label{subsec:res}

\noindent\myparagraph{VOC-COCO Offline Benchmark}. 
We first evaluate the different methods on a typical setting where \id object instances of both train and test data are from VOC dataset and \ood objects are sampled from COCO dataset.
ND results in terms of FPR and AUROC are shown in Table~\ref{tab:initial}. 
Among the methods that do not leverage auxiliary \ood data, VOS~\cite{du2022vos} seems to perform the best. \ours achieves almost $50\%$ reduction rate on the FPR metric compared to VOS~\cite{du2022vos}. Although we train on real \ood data,  it is an evidence on our approach  effectiveness.

\begin{figure*}[t!]
    \centering
    \vspace{-0.7pt}
    \begin{subfigure}[t]{0.3\textwidth}
        \centering
        \includegraphics[width=.88\textwidth]{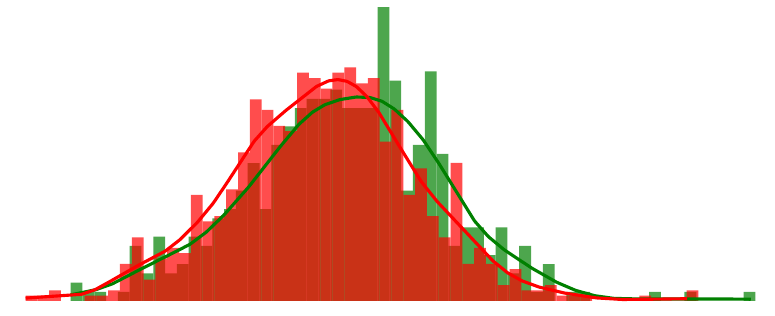}
        \caption{Max Logit}
    \end{subfigure}%
    \hfill
    \begin{subfigure}[t]{0.3\textwidth}
        \centering
        \includegraphics[width=.88\textwidth]{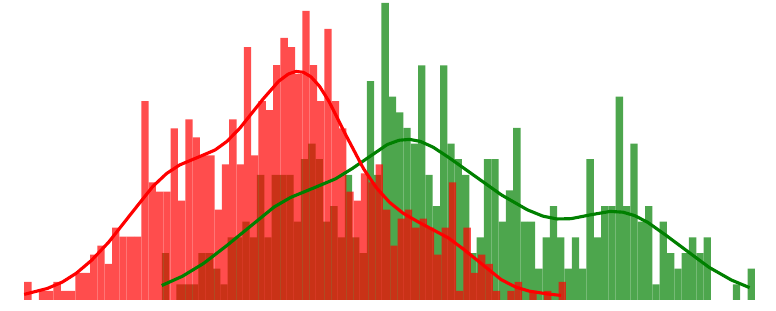}
        \caption{\ours after $S_0$}
    \end{subfigure}
      \hfill
    \begin{subfigure}[t]{0.3\textwidth}
        \centering
        \includegraphics[width=.88\textwidth]{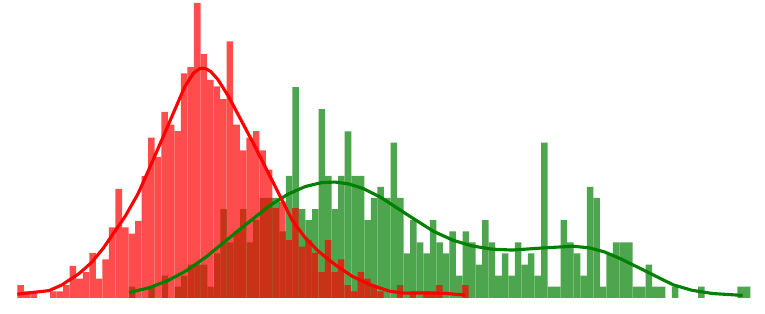}
        \caption{\ours after $S_4$}
    \end{subfigure}
          \hfill
    \begin{subfigure}[t]{0.08\textwidth}
        \centering
        \includegraphics[width=.98\textwidth]{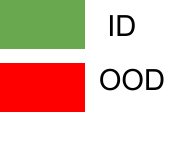}
        
    \end{subfigure}
    \caption{\label{fig:scores} \footnotesize The \id{}-\ood{} scores distributions on $G^{holdout}$ from COCO-OpenImages benchmark predicted by (a) Max Logit, (b) \ours after the initial training session $S_0$, and (c) \ours after $S_4$. Starting with overlapping distribution with Max Logit, \ours separates the \id{}-\ood{} and largely benefits from the received feedback.}
    \vspace{-5mm}
\end{figure*}

\vspace{2mm}
\noindent\myparagraph{VOC-COCO Test-domain benchmark}. 
Here we evaluate the different methodologies on a more realistic yet offline setting. We consider our test-domain benchmark (Sec.~\ref{subsec:tdib} ) where both \id and \ood images and their objects instances are drawn from the same test dataset, COCO, whereas the object detection model is trained on VOC. Results in terms of AUROC and FPR are reported in Table~\ref{tab:initial}.
First, it is clear that this is more challenging  as the  performance of all methods have dropped significantly. Surprisingly, VOS~\cite{du2022vos} is not the best performing among the offline methods. It is in fact on par with MSP and worse than Max Logit. This shows that   VOS  training on synthetic \ood data does not generalize to the case of semantic shift only, \ie, only different classes of objects among \id and \ood,  and that VOS  suffers from the shift between the training dataset and the test environment.
Second, our method \ours still significantly improves the performance, we can see that the FPR drops from $77.6\%$ by Max Logit to $45.37\%$ by \ours with an AUROC increasing by $12\%$.
Third, while both \bce and \ours perform closely, our target is to improve the ND performance as more feedback is received, and to show that it is possible to further enhance the generalization capabilities as more \ood and \id examples become available.

\noindent\myparagraph{Test-Domain Incremental Benchmark.}
Here we evaluate the full pipeline of our setting, \emph{Incremental Object-Based Novelty Detecting with Feedback Loop}, following the test-domain incremental benchmark described in the paragraph~\ref{subsec:tdib}. 
In our previous results, Max Logit is the best performing offline method, hence we use this method performance as a reference to assess the improvement brought by the following training sessions. 
Figure~\ref{fig:incremental_steps}  reports the FPR metric of \bce and \ours after each  training session on $G^{holdout}$. We refer to the Appendix A.4 for full results.\\
On VOC-COCO benchmark (a), \ours performance after the first session is similar to that of \bce. However, as more feedback is received the improvements of \ours become more pronounced. 
By the end of the training sequence, it is clear that \ours benefits greatly from the provided feedback and the incremental updates, with a steady reduction in the FPR. 
The FPR of \ours has dropped from $40.38\%$ to $25.0\%$ compared to only $2.5\%$ drop for \bce after 5 training iterations.%
These results confirm the superiority of our \ours compared to other alternatives. \\
We then consider a larger scale scenario where COCO is used  for training the object detection model and OpenImages is used as a test set with 60 \id object classes and 96 \ood classes.
Figure~\ref{fig:incremental_steps} (b) reports the FPR  of \bce and \ours on $G^{holdout}$. 
Similar trends to what we have seen in VOC-COCO benchmark can be observed here. Both methods start of with similar performance, \ours starts to significantly improve the ND performance by the third incremental step. At the end of the sequence, \ours reduces the FPR from $65.62\%$ to $41.47\%$, compared to $64.88\%$ to $50.18\%$ by \bce. \ours FPR is less than \bce by almost $9\%$ and the AUROC (not shown here) is higher  by $2.5\%$. 
These results suggest the key role of optimizing a representation where \id samples lie close together as opposed to directly optimizing the decision boundaries as in \bce.\\
Additionally, Figure~\ref{fig:scores} shows the distributions of \id-\ood scores by \ours before and after the incremental training compared to Max Logit scores in the COCO-OpenImages benchmark. Starting from completely overlapped distributions by Max Logit (a) associated with high FPR of $93.84$, using only the features extracted from \modd, \ours manages to separate the two distributions after $S_0$ (b) and  successfully learns later from  feedback loop resulting in a near optimal \id-\ood separation (c). 

\begin{figure}[t!]
    \centering
    \includegraphics[width=1\textwidth]{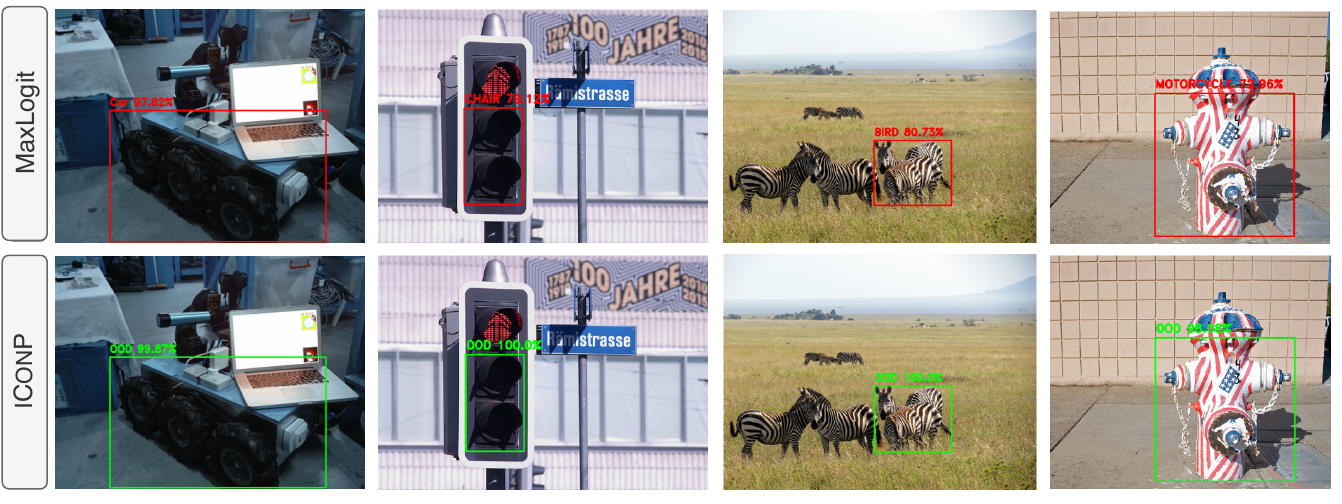}
    \caption{\footnotesize Qualitative results on VOC-COCO benchmark with MaxLogit (above) and \ours (below) after $S_4$. Top row shows images with wrongly detected objects as \id by Max Logit leading to mistaken high confidence classification by \modd. Bottom row, \ours can identify those \ood objects and prevents misleading classification.}
 \label{fig:qualitativeresults}
 \vspace{-2mm}
\end{figure}

\myparagraph{Qualitative Results.}
To inspect how \ood objects are treated by \ours compared to Max Logit, Figure~\ref{fig:qualitativeresults} shows some qualitative examples. Relying  on an offline \ood detection method as Max Logit leads to serious mistakes and as a result   confident yet wrong \id object detections. On the contrary \ours is much robust and can identify those objects as \ood avoiding any mishandling of these cases, leading to safer deployment of the object detection model.

\section{Discussion and Conclusion}
\label{conclusion}
We have shown how existing ND methods struggle when both \id{} and \ood{} objects are sampled from the same testing distribution in Table~\ref{tab:initial}. Training our ND module on an initial set of this testing distribution improves largely the detection performance. When feedback is received and new training sessions are performed, \ours{} exhibits great FPR reduction, \ie{}, from 40.38\% to 25.00\% on VOC-COCO and from 65.62\% to 41.47\% on COCO-OpenImages. This proves that our method can successfully leverage the newly received annotations and increase its robustness. In spite of \ours good results, there is still room for further improvements on how to better learn continuously that we hope future research will address. Our code and benchmarks will be released. 

In conclusion, ensuring the safety of machine learning models in real-life scenarios is critical, particularly in detecting unknown objects and avoiding misleading behaviors. We highlight the importance of utilizing abundant test data and human feedback through inexpensive annotations. Our investigation shows that incorporating feedback on \ood decisions significantly enhances the robustness of pretrained models to \ood data. By updating a small novelty detection module with this feedback, we maintain performance on \id objects and increase the model's trustworthiness in handling \ood data.

{
    \small
    \bibliographystyle{splncs04}
    \bibliography{bibliography}
}

\newpage

\appendix
\section{APPENDIX}
\label{appendix}
\subsection{Implementation Details}
\label{app:implemdet}

For the object detection, we relied on the Detectron2 \cite{wu2019detectron2} library, which provides a variety of pretrained deep learning architectures for the most relevant computer vision tasks at the moment. Specifically, for our experiments on  VOC based benchmarks we use \texttt{faster\_rcnn\_R\_50\_C4}, a Faster R-CNN model with ResNet-50 as backbone trained on Pascal VOC 2007 + VOC 2012, while for those on the COCO based benchmarks we use \texttt{faster\_rcnn\_R\_50\_FPN\_3x}, a Faster R-CNN model with ResNet-50 and Feature Pyramid Network as a backbone trained on COCO 2017.
As input to \mnd{} we use the hidden features predicted in the \emph{roi} heads for each detected object after non-max suppression and objectiveness threshold.

\bce{} is trained using ADAM optimizer with a $lr = 0.0005$ and a Cosine Annealing LR scheduler while employing early stopping (with a window size of 5). 
As for variants (presented in table \ref{tab:regularizers}) with different regularization functions we tested \mnd{} with Dropout with $p=0.5$, Mixup ($x_{mix} = x_i + \lambda \cdot x_j$) at input feature level with $\lambda \sim \mathcal{N}(\mu=0.5,\,\sigma^{2}=0.2)$ as linear interpolation parameter, Manifold Mixup \cite{pmlr-v97-verma19a}, where the Mixup happens at a random hidden layer $L$ at each Mixup step. \\
For \ours we optimized the latent representation with SupCon loss \cite{khosla2020supervised}, applied at the penultimate layer of the model, and BCE applied, in a disjoint fashion, to the last layer of the model, intended as a 1D projection to the \emph{score space}. To alleviate any possible instability of SupCon, we initialized the weights of the model using Xavier initialization and a warmup for the first 10 epochs, varying the $lr$ from 0.01 to 0.005. We found out the SGD optimizer to be a better match than Adam along with a Cosine Annealing LR scheduler, where the \emph{lr} is updated before each epoch. 
Differently from the model trained with BCE, we keep a fixed  number of epochs $e=25$, since SupCon generally requires more number of epochs before obtaining a good features representation. Once a model has been trained on the first split $G_0$, it is incrementally updated employing the subsequent splits. During this process we reduce the learning rate by a factor of 10 for all our variants and we set a fixed number of epochs for both the \bce{} and \ours{} based ones, respectively 5 epochs and 15 epochs, after which each model seems to converge.

In our benchmarks, VOC-COCO test domain and COCO-OpenImages incremental, we used respectively splits generated from COCO and OpenImages datasets.
Starting from 80 classes in total in COCO dataset we first filter the 20 classes contained also in VOC as our \id{} data and we keep the remaining 60 classes as \ood{} data. We then generate $G_0$ by filtering  30 out of 60 \ood{} classes and the remaining 30 have been evenly distributed to generate the latter 5 splits ($\{G_1,~G_2,~G_3,~G_4,~G^{holdout}\}$). 
Differently from COCO, given the large amount classes in OpenImages, after having extracted 60 (out of the 80) COCO \id{} classes; we filter out 96 of the remaining 500+ \ood{} classes to form our \ood{} dataset. We chose to keep only \emph{highly} represented classes with at least 5000 occurrences out of 15M bbox annotations, still maintaining the classical long tail shape distribution of \ood{} classes. 
Similar as before, we keep half of the \ood{} classes as our $S_0$ (48 classes), and the remaining 48 classes form our incremental sessions i.e., 10, 10, 10, 9, 9 classes respectively.

\subsection{Computational Overhead}
\label{compover}

Since we rely on the features extracted from the frozen object detection model, the computational overhead of our novelty detection methodology appears negligible. Specifically, at inference time of our Novelty Detection module (~150k parameters) represents the 1\% of the average time required for the object detection inference. Similarly, the resources for one training session is minimal in time and compute, as it takes ~10 seconds of training time for a few thousands of samples (on a budget gpu, i.e. Nvidia gtx 2080, with 5 training epochs, 8 GB of vram and a batch size of 512 samples). This illustrates the efficiency of our approach and that increasing a model robustness to \ood{} samples can be quite cheap in terms of annotations and compute as opposed to updating the main object detection model performance.

\subsection{Can \bce generalize better when combined with different regularization functions?}

In the experiments section, we have shown that \bce exhibits poor performance improvements as more feedback is received and we have motivated our choice of loss with the fact that directly optimizing decision boundaries by binary cross entropy doesn't generalize well to unseen \ood classes and doesn't allow enough flexibility when new samples and \ood classes are received. 

To further confirm the superiority of our design choice and motivation, we combine  \bce with different regularization techniques and inspect whether any improvements in generalization can be observed compared to \ours.  We consider different regularization techniques namely, dropout~\cite{dropout}, Mixup~\cite{zhang2017mixup} and manifold mixup~\cite{pmlr-v97-verma19a}. Table~\ref{tab:regularizers} reports the FPR and AUROC of \bce with each regularizer and the combination of multiple regularizers in our test-domain incremental  benchmark (VOC-COCO). 
It can be seen that some regularizers provide  improvements at different training sessions but these improvements are not consistent and the overall performance is significantly inferior to that of our method  \ours{}. Moreover, it is interesting to notice that, while \bce{} trained models consistently show better initial performance, \ours{} outperforms all \bce variants as the training progresses, suggesting that regularization is not sufficient to improve \ood detection performance as training session are carried.  This proves that  \ours ability of steady performance improvements is not trivial.

\begin{table}
\caption{\label{tab:regularizers}\footnotesize Novelty detection performance of \bce{} with different regularization mechanisms and \ours{} on the test-domain incremental benchmark (VOC-COCO).}
\resizebox{0.98\textwidth}{!}{ 
\centering
\setlength{\tabcolsep}{2.8pt}
\begin{tabular}{@{}lcccccccccccc@{}} \toprule

\multirow{2}{*}{\textbf{Method}} & \multicolumn{2}{c}{$S_0$} & \multicolumn{2}{c}{$S_{1}$} & \multicolumn{2}{c}{$S_{2}$} & \multicolumn{2}{c}{$S_3$} & \multicolumn{2}{c}{$S_4$}\\

& FPR@95$(\downarrow$) & AUROC$(\uparrow$) & FPR@95$(\downarrow$) & AUROC$(\uparrow$) & FPR@95$(\downarrow$) & AUROC$(\uparrow$) & FPR@95$(\downarrow$) & AUROC$(\uparrow$) & FPR@95$(\downarrow$) & AUROC$(\uparrow$) \\

\midrule

\bce{} & 37.50 & 91.37 & 35.57 & 91.65 & 42.30 & 90.98 & 37.50 & 92.22 & 35.57 & 92.65 \\
\midrule
\bce{} +  Dropout \cite{dropout} & 35.9 & 91.88 & 34.51 & 92.19 & 38.25 & 91.67 & 33.97 & 92.96 & 32.26 & 93.46 \\ 
\midrule
\bce{} +  Mixup \cite{zhang2017mixup} & 43.91 & 89.35 & 40.17 & 91.59 & 41.03 & 91.07 & 36.43 & 92.43 & 34.83 & 92.89 \\ 
\midrule
\bce{} + Manifold Mixup \cite{pmlr-v97-verma19a} & 38.78 & 89.62 & 38.46 & 89.58 & 38.46 & 89.02 & 38.03 & 89.84 & 37.39 & 90.33 \\ 
\midrule
\bce{} + Dropout + Mixup & \textbf{35.58} & 92.58 & \textbf{33.65} & 92.57 & 34.72 & 91.99 & 31.52 & 93.17 & 31.2 & 93.67 \\
\midrule
\bce{} + Dropout + Manifold Mixup & 38.78 & 90.29 & 39.53 & 90.17 & 39.96 & 89.56 & 39.74 & 90.28 & 38.03 & 90.75 \\ 
\midrule
\ours{} & 40.38 & \textbf{93.11} & 38.46 & \textbf{93.89} & \textbf{34.61} & \textbf{93.75} & \textbf{26.92} & \textbf{94.33} & \textbf{25.00} & \textbf{94.19} \\

  \bottomrule
\end{tabular}}

\end{table}

\subsection{Detailed Results}~\label{sec:fullresults}

Due to the space limits, in the main experiments we only show FPR metric in a plot for our test-domain incremental  benchmark. Here we report the detailed numbers of both FPR and AUROC after each training session. Additionally we expand our evaluation protocol to evaluate on two holdout sets: 1) a holdout set of unseen ood classes $G^{holdout}$ (as in the main paper) and 2) a holdout set composed of both unseen and encountered ood classes of $G_{0, 1, 2, 3, 4}$ as it is natural to re-encounter some \ood classes in practice, and it is important to see the ND performance on \textit{ unseen} samples of those previously encountered \ood classes.
Tables \ref{tab:incrementalcococomp} and \ref{tab:incrementaloimgcomp} report the FPR and AUROC for \bce and \ours after each training session on VOC-COCO and COCO-OpenImages respectively.

On both holdout sets and and both incremental benchmarks, \ours shows steady improvements and superior performance to that of \bce. 

\begin{table}[h]
\caption{\label{tab:incrementalcococomp}\footnotesize $ND$ performance on {VOC-COCO incremental test-domain benchmark}. As training sessions progress {\ours} benefits greatly from the feedback and reduce significantly the FPR while improving the AUROC.}
\resizebox{0.98\textwidth}{!}{ 
\centering
\setlength{\tabcolsep}{2.8pt}
\begin{tabular}{@{}lcccccccccccc@{}} \toprule

\multirow{2}{*}{\textbf{Method}} & \multirow{2}{*}{\textbf{Eval}} & \multicolumn{2}{c}{$S_0$} & \multicolumn{2}{c}{$S_{1}$} & \multicolumn{2}{c}{$S_{2}$} & \multicolumn{2}{c}{$S_3$} & \multicolumn{2}{c}{$S_4$}\\

&& FPR@95$(\downarrow$) & AUROC$(\uparrow$) & FPR@95$(\downarrow$) & AUROC$(\uparrow$) & FPR@95$(\downarrow$) & AUROC$(\uparrow$) & FPR@95$(\downarrow$) & AUROC$(\uparrow$) & FPR@95$(\downarrow$) & AUROC$(\uparrow$) \\
\midrule
\multirow{2}{*}{\bce}   & $G_{0, 1, 2, 3, 4}$ & \textbf{47.37} & \textbf{91.24} & \textbf{35.72} & 92.42 & 29.12 & 94.01 & 26.79 & 94.92 & 24.85 & 95.28 \\
                        & $G^{holdout}$ & \textbf{37.50} & 91.37 & \textbf{35.57} & 91.65 & 42.30 & 90.98 & 37.50 & 92.22 & 35.57 & 92.65 \\
\midrule
\multirow{2}{*}{\ours}  & $G_{0, 1, 2, 3, 4}$ & 47.76 & 90.61 & 39.41 & \textbf{92.57} & \textbf{21.16} & \textbf{95.44} & \textbf{19.22} & \textbf{96.41} & \textbf{17.86} & \textbf{96.54} \\
                        & $G^{holdout}$  & 40.38 & \textbf{93.11} & 38.46 & \textbf{93.89} & \textbf{34.61} & \textbf{93.75} & \textbf{26.92} & \textbf{94.33} & \textbf{25.00} & \textbf{94.19} \\
\bottomrule
\end{tabular}
}

\end{table}

\begin{table}[h]
\caption{\label{tab:incrementaloimgcomp}\footnotesize ND  performance on {COCO-OpenImages incremental test-domain benchmark}.  As training sessions progress {\ours} benefits greatly from the feedback and reduce significantly the FPR while improving the AUROC.}
\resizebox{0.98\textwidth}{!}{ 
\centering
\setlength{\tabcolsep}{2.8pt}
\begin{tabular}{@{}lcccccccccccc@{}} \toprule

\multirow{2}{*}{\textbf{Method}} & \multirow{2}{*}{\textbf{Eval}} & \multicolumn{2}{c}{$S_0$} & \multicolumn{2}{c}{$S_{1}$} & \multicolumn{2}{c}{$S_{2}$} & \multicolumn{2}{c}{$S_3$} & \multicolumn{2}{c}{$S_4$}\\

&& FPR@95$(\downarrow$) & AUROC$(\uparrow$) & FPR@95$(\downarrow$) & AUROC$(\uparrow$) & FPR@95$(\downarrow$) & AUROC$(\uparrow$) & FPR@95$(\downarrow$) & AUROC$(\uparrow$) & FPR@95$(\downarrow$) & AUROC$(\uparrow$) \\
\midrule
\multirow{2}{*}{\bce}   & $G_{0, 1, 2, 3, 4}$ & \textbf{75.77} & 81.14 & 74.40 & \textbf{82.68} & 66.52 & 85.93 & 62.37 & 87.00 & 53.53 & 88.83 \\
                        & $G^{holdout}$ & \textbf{64.88} & 82.45 & 63.76 & \textbf{83.98} & 59.40 & 86.74 & 57.53 & 87.48 & 50.18 & 89.10 \\
\midrule
\multirow{2}{*}{\ours}  & $G_{0, 1, 2, 3, 4}$ & 77.00 & \textbf{81.38} & \textbf{74.29} & 82.56 & \textbf{61.46} & \textbf{87.73} & \textbf{57.17} & \textbf{88.83} & \textbf{47.88} & \textbf{90.75} \\
                          & $G^{holdout}$ & 65.62 & \textbf{83.23} & \textbf{63.51} & 83.93 & \textbf{52.42} & \textbf{88.68} & \textbf{48.81} & \textbf{89.49} & \textbf{41.47} & \textbf{91.58} \\
\bottomrule
\end{tabular}
}

\end{table}

\end{document}